%% file: 0_main.tex
\documentclass[10pt,twocolumn,letterpaper]{article}

\usepackage{cvpr}

\usepackage{graphicx}
\usepackage{amsmath}
\usepackage{amssymb}
\usepackage{booktabs}

\usepackage{verbatim}

\usepackage{color}

\usepackage{marvosym}

\usepackage[pagebackref,breaklinks,colorlinks]{hyperref}

\usepackage[capitalize]{cleveref}
\crefname{section}{Sec.}{Secs.}
\Crefname{section}{Section}{Sections}
\Crefname{table}{Table}{Tables}
\crefname{table}{Tab.}{Tabs.}

\def\cvprPaperID{6788} 
\def\confName{CVPR}
\def\confYear{2023}

\newcommand{\datasetname}{{\sc Describe3D }}

\begin{document}


\title{High-Fidelity 3D Face Generation from Natural Language Descriptions}

\author{Menghua Wu, Hao Zhu$\textsuperscript{\Letter}$, Linjia Huang, Yiyu Zhuang, Yuanxun Lu, Xun Cao\\ \\
Nanjing University, Nanjing, China
}

\maketitle

\begin{abstract}
Synthesizing high-quality 3D face models from natural language descriptions is very valuable for many applications, including avatar creation, virtual reality, and telepresence. However, little research ever tapped into this task. 
We argue the major obstacle lies in 1) the lack of high-quality 3D face data with descriptive text annotation, and 2) the complex mapping relationship between descriptive language space and shape/appearance space.
To solve these problems, we build \datasetname dataset, the first large-scale dataset with fine-grained text descriptions for text-to-3D face generation task.
Then we propose a two-stage framework to first generate a 3D face that matches the concrete descriptions, then optimize the parameters in the 3D shape and texture space with abstract description to refine the 3D face model. 
Extensive experimental results show that our method can produce a faithful 3D face that conforms to the input descriptions with higher accuracy and quality than previous methods. 
The code and \datasetname dataset are released at \url{https://github.com/zhuhao-nju/describe3d}.
\end{abstract}

\input{1_intro.tex}

\input{2_related.tex}

\input{3_method.tex}

\input{4_exp.tex}

\input{5_con.tex}

\input{6_supple.tex}

\clearpage

{\small
\bibliographystyle{ieee_fullname}
\bibliography{egbib}
}

\end{document}

%% file: 1_intro.tex
\section{Introduction}
\label{sec:intro}

3D faces are highly required in many cutting-edge technologies like digital humans, telepresence, and movie special effects, while creating a high-fidelity 3D face is very complex and requires vast time from an experienced modeler. Recently, many efforts are devoted to the synthesis of text-to-image and image-to-3D, but they lack the ability to synthesize 3D faces given an abstract description. However, there is still no reliable solution to synthesize high-quality 3D faces from descriptive texts in natural language. 

\begin{figure}
    \centering
    \includegraphics[width=1.0\linewidth]{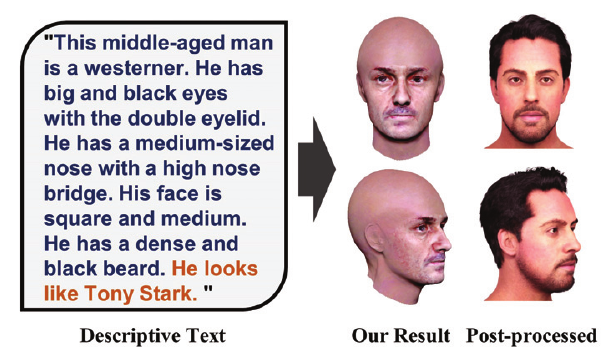}
    \vspace{-0.2in}
    \caption{Given a text describing the appearance (\textbf{\emph{left}}), our method can synthesize high-quality 3D faces (\textbf{\emph{middle}}) containing 3D mesh and textures. The resulting model can be easily processed into a rigged face with hair and accessories (\textbf{\emph{right}}). The dark blue texts indicate concrete descriptions and the brown texts indicate abstract descriptions, and similarly hereinafter. 
    }
    \vspace{-0.1in}
    \label{fig:title}
\end{figure}

We consider the difficulties of synthesizing high-quality 3D face models from natural language descriptions lie in two folds. 
Firstly, there is still no available fine-grained dataset that contains 3D face models and corresponding text descriptions in the research community, which is crucial for training learning-based 3D generators. Beyond that, it is difficult to leverage massive 2D Internet images to learn high-quality text-to-3D mapping. 
Secondly, cross-modal mapping from texts to 3D models is non-trivial. Though the progress made in text-to-image synthesis is instructive, the problem of mapping texts to 3D faces is even more challenging due to the complexity of 3D representation. 

In this work, we aim at tackling the task of high-fidelity 3D face generation from natural text descriptions from the above two perspectives.
We first build a 3D-face-text dataset (named {\sc Describe3D}), which contains $1,627$ high-quality 3D faces from HeadSpace dataset~\cite{dai2020statistical} and FaceScape dataset~\cite{yang2020facescape, zhu2021facescape}, and fine-grained manually-labeled facial features. The provided annotations include 25 facial attributes, each of which contains 3 to 8 options describing the facial feature.  Our dataset covers various races and ages and is delicate in 3D shape and texture.
We then propose a two-stage synthesis pipeline, which consists of a concrete synthesis stage mapping the text space to the 3D shape and texture space, and an abstract synthesis stage refining the 3D face with a prompt learning strategy. 
The mapping for different facial features is disentangled and the diversity of the generative model can be controlled by the additional input of random seeds. 
As shown in Figure~\ref{fig:title}, our proposed model can take any word description or combination of phrases as input, and then generate an output of a fine-textured 3D face with appearances matching the description.
Extensive experiments further validate that the concrete synthesis can generate a detailed 3D face that matches the fine-grained descriptive texts well, and the abstract synthesis enables the network to synthesize abstract features like ``wearing makeup'' or ``looks like Tony Stark''.

In summary, our contributions are as follows:
\vspace{-.2cm}
\begin{itemize}
    \setlength\itemsep{-.1cm}
    \item We explore a new topic of constructing a high-quality 3D face model from natural descriptive texts and propose a baseline method to achieve such a goal. 
    \item A new dataset - \datasetname is established with detailed 3D faces and corresponding fine-grained descriptive annotations. The dataset will be released to the public for research purposes.
    \item The reliable mapping from the text embedding space to the 3D face parametric space is learned by introducing the descriptive code space as an intermediary, which forms the core of our concrete synthesis module. Region-specific triplet loss and weighted $\ell_1$ loss further boost the performance.
    \item Abstract learning based on CLIP is introduced to further optimize the parametric 3D face, enabling our results to conform with abstract descriptions.
\end{itemize}

%% file: 2_related.tex
\section{Related Work}
\label{sec:related}

To the best of our knowledge, work that directly studies text-to-3D-face generation is quite limited.
In this section, we review three relevant topics and discuss the connections along with differences with our proposed task and method.

\noindent \textbf{Text-to-shape.} 
Chen \etal~\cite{chen2018text2shape} proposed to generate colored 3D shapes from natural language by learning implicit cross-modal connections between language and physical properties of 3D shapes. 
In further research, Liu \etal~\cite{liu2022towards} proposed to decouple the shape and color predictions for learning features in both texts and shapes and propose the word-level spatial transformer to correlate word features from text with spatial features from shape. 
In several subsequent studies~\cite{michel2022text2mesh, hong2022avatarclip, canfes2022text}, 
CLIP~\cite{radford2021learning} played an important role which is a large pre-trained vision-language model, and prompt learning is leveraged to harness the powerful representation of the CLIP model.
Jain \etal~\cite{jain2022zero} proposed to combine neural rendering with multi-modal image and text representations to synthesize diverse 3D objects from natural language descriptions, and Poole \etal~\cite{poole2022dreamfusion} further leverage a pre-trained 2D text-to-image diffusion model and NeRF~\cite{mildenhall2021nerf} to perform text-to-3D synthesis with more plausible synthesis.

It is worth noting that among the above researches, only Canfes \etal~\cite{canfes2022text} attempted to generate a 3D face, but their model relies on an unconstrained initial 3D face and only work for short phrases. Leveraging facial priors to achieve fine-grained and high-quality 3D face generation from texts in natural language is still an open problem.

\noindent  \textbf{Text-to-image.}
The study of text-to-2D-image started earlier than that of text-to-3D-shape, most of which are based on the generative adversarial network (GAN) \cite{goodfellow2020generative}.
In earlier research, Reed \etal~\cite{reed2016generative, reed2016learning} developed a GAN-based deep architecture to generate plausible images of birds and flowers from detailed text descriptions.
Zhang \etal~\cite{zhang2017stackgan, zhang2018stackgan++} proposed stacked generative adversarial networks, which leverage a sketch-refinement process to enhance the resolution of text-driven image generation.
Dong \etal~\cite{dong2017semantic} proposed a way of synthesizing realistic images given a source image and natural language description and verifying its effectiveness on birds and flowers datasets.
In recent years, GAN-based text-to-image methods have come a long way. The progresses include attention-driven multi-stage refinement~\cite{xu2018attngan}, hierarchical semantic inferring layout~\cite{hong2018inferring}, global-local attentive and semantic-preserving framework~\cite{qiao2019mirrorgan}, semantic decomposing~\cite{yin2019semantics}, StyleGAN inversion module~\cite{xia2021tedigan}. Sun \etal~\cite{sun2022anyface} proposed the diverse triplet loss to learn an accurate mapping from the embedding space of CLIP~\cite{radford2021learning} to parametric space of styleGAN~\cite{karras2019style}. 
Very recently, diffusion model~\cite{ho2020denoising} shows powerful performance in this task~\cite{dhariwal2021diffusion} and synthesizes impressive images reflecting a high-level understanding of the input description. 

The above research works have an enlightening effect on the research of synthesizing a 3D face from descriptive texts, such as the use of the CLIP model, but the two tasks are still very different. Firstly, the representation of 3D faces is much more complex than that of 2D images. Secondly, unlike 2D images that can be easily obtained from the Internet in large quantities, there are very few available 3D face models. These factors determine that text-to-image methods cannot be directly applied to the task of text-to-3D.

\begin{figure*}[th]
    \centering
    \includegraphics[width=1.0\linewidth]{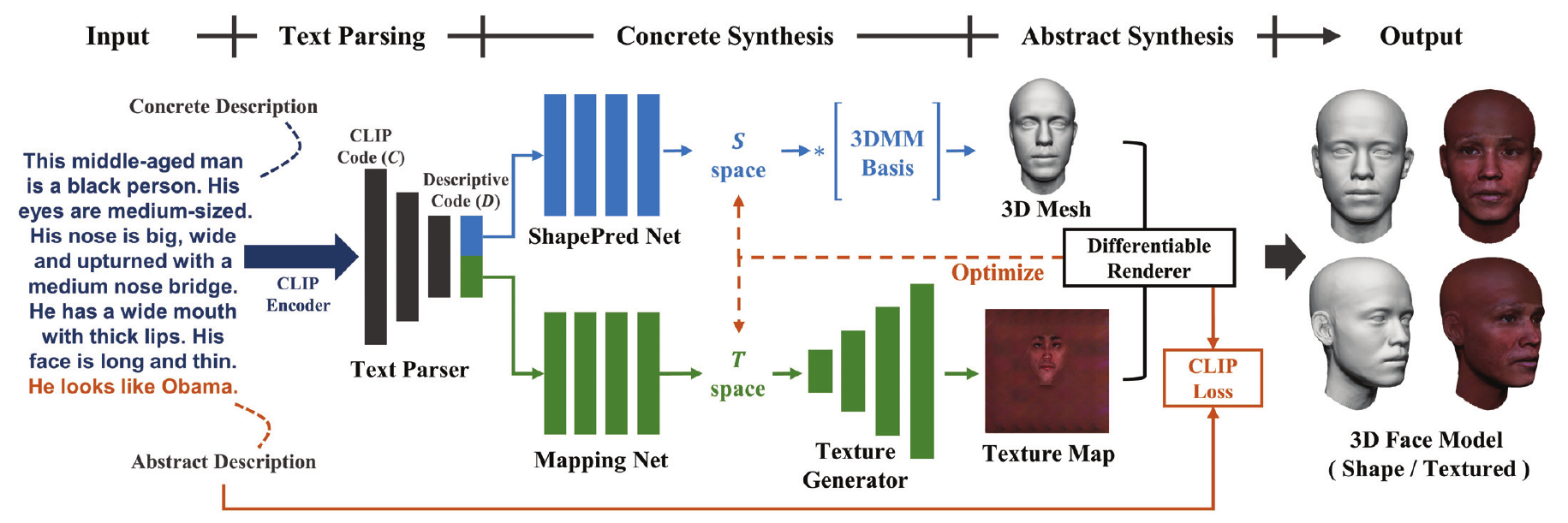}
    \vspace{-0.3in}
    \caption{The overall pipeline consists of three stages: text parsing (Section~\ref{sec:text}), concrete synthesis (Section~\ref{sec:concrete}), and abstract synthesis (Section~\ref{sec:abstract}).
    The dark blue texts indicate concrete descriptions and the brown texts indicate abstract descriptions, and similarly hereinafter.
    }
    \vspace{-0.1in}
    \label{fig:pipeline}
\end{figure*}

\noindent \textbf{3D Face Generation.} 
Early in 1999, Blanz \etal~\cite{blanz1999morphable} propose a 3D morphable model (3DMM) that is a statistical model built upon a set of 3D faces. 
Since then, 3DMM has evolved considerably, and we recommend reading Egger \etal's survey~\cite{egger20203d} for a comprehensive understanding of these advances. With the breakthrough development of deep learning algorithms, 3DMM is widely used in the task of recovering 3D faces from single image~\cite{zhu2016face, guo2020towards, yang2020facescape, sanyal2019learning, feng2021learning} or multiple images~\cite{bai2020deep, xiao2022detailed}, but the research on generating face models from natural text descriptions is very limited. In recent years, some new attempts have been made to use implicit models such as neural radiation field~\cite{lombardi2021mixture, hong2022headnerf, zhuang2021mofanerf}, signed distance field (SDF)~\cite{yenamandra2021i3dmm, guo2023rafare} and other implicit representations\cite{zheng2022avatar, zheng2022imface} to represent 3D faces. 

3D face generation is one of the key components of our task and defines the parametric space for 3D faces, while our work further studies the mapping problem from the text description space to the parametric space of 3D faces.

%% file: 3_method.tex
\section{Method}
\label{sec:method}

In this work, we aim to synthesize a high-quality and faithful 3D head from natural text descriptions. 
To this end, a three-stage learning-based pipeline is proposed as shown in Figure~\ref{fig:pipeline}.
The text encoder (Section~\ref{sec:text}) first parses the input natural texts and generates a descriptive vector, which is then fed into the module of concrete synthesis (Section~\ref{sec:concrete}) to predict 3D shape and texture separately. 
The generated 3D shape and texture are then optimized by abstract synthesis (Section~\ref{sec:abstract}), then the result 3D face is generated. Our results can be easily processed into a riggable 3D face with full assets. We now explain these sub-modules in detail.

\subsection{{\sc Describe3D} Dataset}
\label{sec:dataset}

\begin{figure}
    \centering
    \includegraphics[width=0.9\linewidth]{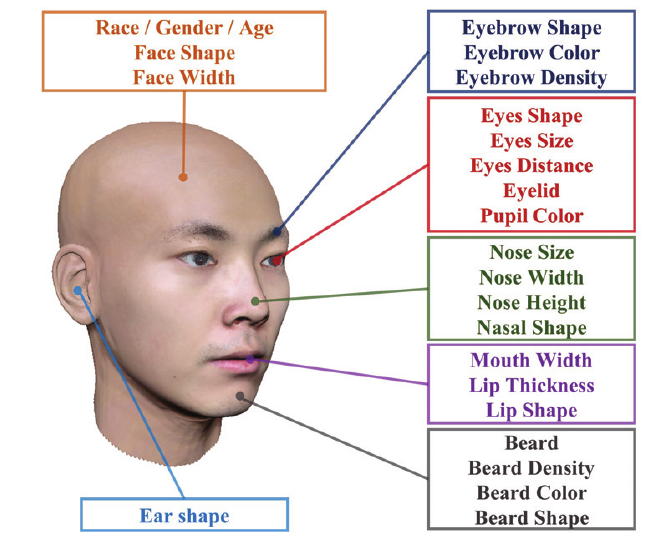}
    \vspace{-0.0in}
    \caption{Our annotations of the 3D faces contain $25$ single-choice questions regarding to the attributes shown above and a free-style text description. These attributes are categorized into shape-related, color-related, and general-related attributes. A complete questionnaire will be provided in the supplementary material. 
    }
    \vspace{-0.1in}
    \label{fig:anno}
\end{figure}

To establish an accurate mapping from natural language to 3D faces, we first need pairs of the 3D model and its matching text description.
However, to the best of our knowledge, there is no 3D face model dataset with detailed textual descriptions available. In this work, we build the \textit{first} fine-grained descriptive 3D face dataset (referred to as {\sc Describe3D} dataset) to train our text-to-3D-face model.

Our dataset contains $1,627$ 3D face models collected from HeadSpace~\cite{dai2020statistical} and FaceScape~\cite{yang2020facescape, zhu2021facescape} datasets, covering the four major races: Mongoloid, Caucasoid, Negroid, Australoid, and with the range of ages from $16$ to $69$. 
We process the raw scanned 3D faces from HeadSpace and FaceScape to uniform their mesh topology.
For 3D shape representation, we align all 3D faces into a canonical space with Procrustes analysis~\cite{gower1975generalized} and non-rigid iterative closest point (NICP) algorithm~\cite{amberg2007optimal}.  These aligned 3D faces are assigned with a uniform mesh topologically containing $26,369$ vertices and $52,536$ triangle faces. 
For texture representation, we align all texture maps into a uniform UV coordinate that is attached to the uniform mesh topology.

We then manually annotate these 3D faces to obtain detailed facial shapes and appearance features. 
As shown in Figure~\ref{fig:anno}, our annotations of the 3D faces contain $25$ labels from single-choice questions and a free-style text description, covering features including facial shape, appearance, and free-style descriptions. 
Then, we generate a concrete descriptive text by filling all features into multiple pre-designed sentence templates, such as ``His {\tt [eyes]} are {\tt [medium-sized]}'' and ``He has {\tt [wide mouth with thick lips]}'' (an example shown in Figure~\ref{fig:pipeline}). A detailed list of sentence patterns will be shown in the supplementary material.
These generated sentences are finally combined with the collected text written by human annotators to form a complete description of a human face, which we refer to as \textit{concrete descriptions} (as opposed to \textit{abstract descriptions} to be defined in \Cref{sec:abstract}). The average length of concrete descriptions is $79$ tokens.

\subsection{Text Parser}
\label{sec:text}

The text parser aims at encoding the input natural language into a descriptive code $d$ that can be mapped into the 3D facial model space.
In this work, we adopt CLIP~\cite{radford2021learning}, a large language–image pre-training model, to encode the description texts into a CLIP embedding. 
In pilot studies, we observed that directly predicting 3D faces from such embedding did badly in mapping performance, partly because of the high complexity of the text descriptions.
We thus propose to first predict a descriptive code derived from our labeled data from the CLIP embedding, and then synthesize the 3D faces from such descriptive code. 
Specifically, the descriptive code is a $p \times q$ matrix, with $p$ rows representing $p$ different annotated facial attributes, and each column is an $q$-dimension one-hot vector describing this attribute.
The motivation behind such design is simple -- the introduction of the descriptive code decomposes a complex mapping task into two simpler tasks: to predict the descriptive code from the text and to synthesize 3D faces from the descriptive code. 

Our text parser is an 8-layer MLP, which takes the CLIP embedding as input and predicts the descriptive code. Given the predicted code $\hat{y}$ and ground-truth $y$, the loss function to train the text parser is formulated as:
\begin{align}
    {{L}_{parse} = -\frac{1}{p} \sum\limits_{i=1}^p\sum\limits_{j=1}^q y_{ij}\log\text{softmax}(\hat{y}_{ij})},
\end{align}
where $i$ is the index of the annotated facial attributes, and $j$ is the index of the feature option to describe this attribute. 
As 3D registration loses most features about the ear in the \datasetname dataset, the descriptive code doesn't contain the annotation of ear shape. So we set $p=24$ and $q=8$ in all the experiments.

\subsection{Concrete Synthesis}
\label{sec:concrete}

The network of concrete synthesis takes the predicted descriptive code as input and aims at generating a set of diverse 3D faces that faithfully match the concrete text descriptions.
Considering that a 3D face model contains 3D shapes and textures, we first separate the descriptive code $d$ into shape-related code $d_S$ and texture-related code $d_T$ according to our annotation, then use two sub-networks to synthesize 3D shapes and textures, respectively.

\noindent \textbf{Shape Generation Network.}
We leverage a 3D morphable model (3DMM) to represent the 3D facial shape in a $S$-space, and an MLP is used to predict 3DMM parameters $s \in V$ from the shape-related descriptive code $d_S$, referred to as ShapePred Net in Figure~\ref{fig:pipeline}.
Other than predicting 3DMM parameters, another approach to generate 3D shapes is to directly predict a 3D polygon mesh~\cite{feng2019meshnet} or a position map~\cite{gecer2020synthesizing}. In essence, the introduction of 3DMM is equivalent to converting large-scale 3D shapes into low-dimensional parametric space, which provides a strong prior to reducing the difficulty of the shape generation task. 
Through the experiments (\ref{sec:ablation}), we found that predicting 3DMM parameters leads to more accurate mapping than directly predicting position maps in our task.

Following FaceScape~\cite{yang2020facescape, zhu2021facescape}, we generate the 3DMM model from the 3D polygon mesh models in the training set with Principle Components Analysis (PCA)~\cite{wold1987principal}. 
Specifically, given $m$ facial mesh models and each of which contains $n$ vertices, a $m \times n$ tensor is built representing all these vertices in the training set. 
We use Tucker decomposition~\cite{tucker1966some} to decompose the $m \times n$ tensor to a small PCA basis matrix $B$ and a lower $m'$-dimensional factor representing facial identity. A new set of vertices $v$ representing 3D face shape can be generated given an arbitrary 3DMM parameter $s$ as: 
\begin{align}
v = B \times \emph{s}.
\end{align}

In this way, large-scale data of 3D facial shapes are mapped into an $m'$-dimensional parameter space, referred to as $S$-space. 
In all our experiments, we set $m=1,627$, $m'=300$, and $n=26,369$.

In this work, we experiment with the following two types of losses to train the ShapePred Net: weighted $\ell_1$ loss and region-specific triplet (RST) loss.

\noindent \textbf{Weighted $\ell_1$ loss.} Through a differentiable 3DMM mapping module, the predicted 3DMM parameters can be transformed into the 3D positions of the vertices. We found that applying $\ell_1$ loss directly to all vertices resulted in an overall average result. 
We use a weighted mask similar to PRNet~\cite{feng2018joint} to calculate the loss for different regions. The weighted $\ell_1$ loss function is formulated as:
\begin{align}
{L}_{w \ell_1} = \sum\limits_i{\alpha}_i \times \Vert{\hat{v_i} - v_i\Vert_1},
\end{align}
where $v_i$ represents the vertices of $i$-th region, and $\alpha_i$ represents the corresponding weight. Here we divide the whole head mesh model into four regions: (1) $68$ facial landmarks; (2) eyes, nose, and mouth; (3) the other facial regions; and (4) the back of the head with ears. The weights for these regions are set as $16:4:3:0$. 

\noindent \textbf{Region-specific Triplet (RST) Loss.} 
To enhance the diversity of the generated 3D shape, we propose RST loss to train the 3DMM regressor. Triplet loss was firstly proposed in FaceNet~\cite{schroff2015facenet} and widely used in the task of face recognition, then was introduced into the task of image generation~\cite{wang2021age, sun2022anyface}. 
The key idea behind this is to make the difference between prediction and positive examples minor, and the difference between prediction and negative examples greater. 

\begin{figure}
    \centering
    \includegraphics[width=1.0\linewidth]{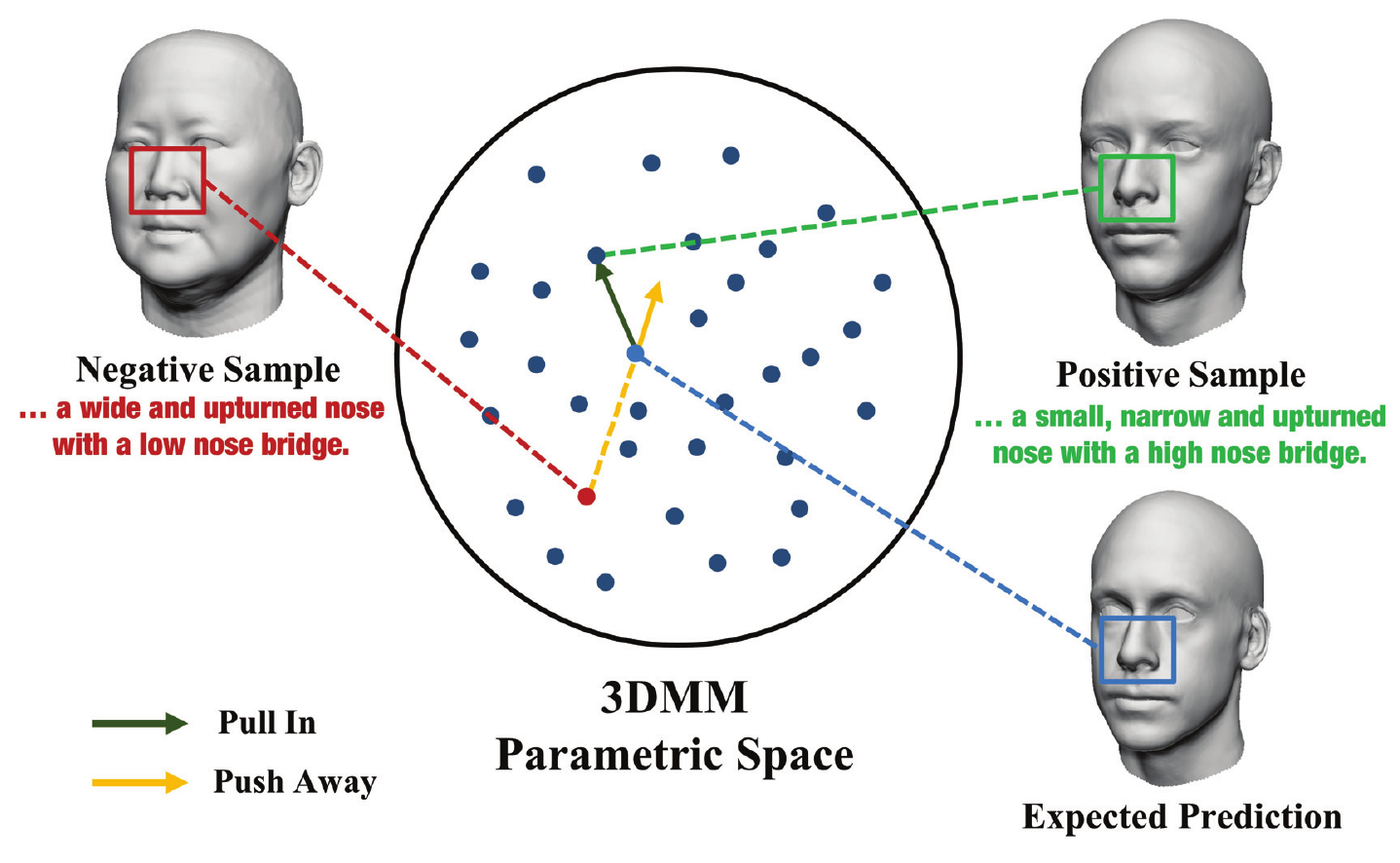}
    \vspace{-0.2in}
    \caption{
    Region-Specific Triplet loss (RST loss). For a specific region like the nose, RST loss pushes the prediction away from the negative sample and close to the positive sample. 
    }
    \vspace{-0.1in}
    \label{fig:dtloss}
\end{figure}

Different from previous works that measure the difference of the samples in the parametric space, we propose to measure the difference with mean Euclidean distance and apply weights for different regions. 
Specifically, we divide the human face into eyes, nose, mouth and others, and treat them separately in the training phase. 
As shown in Figure~\ref{fig:dtloss}, in each training iteration, we randomly select positive-negative pairs for a random region and compute RST loss, which is formulated as:
\begin{align}
    {L}_{RST} = \max(\Vert{\hat{v_i} - v_i}\Vert_1 - \Vert \hat{v_i} -  v_i^{*}\Vert_1 + m_i , 0) \cdot {\lambda}_i,
\end{align}
where $\hat{v_i}$ is the predicted vertices of $i$-th region, $v_i$ is the corresponding ground-truth, and $v_i^*$ is its counter example. $m_i$ and ${\lambda}_i$ represent corresponding region margin and weight respectively.

\noindent \textbf{Texture Generation Network.}
We represent the color of 3D faces with UV texture maps that are attached to the triangle mesh generated by our 3DMM. As shown in Figure~\ref{fig:pipeline}, we adopt a mapping net to map the shape-related descriptive code $d_S$ into a 3DMM code $s$, and a texture generator network to synthesize a UV texture map from the parameter in $T$ space. Here we use StyleGAN2\cite{Karras2019stylegan2, richardson2021encoding, tov2021designing} as the backbone, which is an alternative generator architecture for generative adversarial networks. The input of StyleGAN is a random latent code together with a condition code representing facial features, then these codes are mapped into a $W$ space where different facial features are disentangled, and the 2D images are synthesized from $w \in W$ by a convolutional neural network. In our implementation, our mapping net, texture generator, and $T$ space are corresponding to the mapping network, synthesis network, and $W$ space of StyleGAN, respectively. In the training phase, the StyleGAN2 is re-trained with the UV texture maps in our \datasetname dataset as images, and the descriptive code $d_T$ as the condition input. The loss function and hyper-parameters are the same as the StyleGAN2.

\begin{figure}
    \centering
    \includegraphics[width=1.0\linewidth]{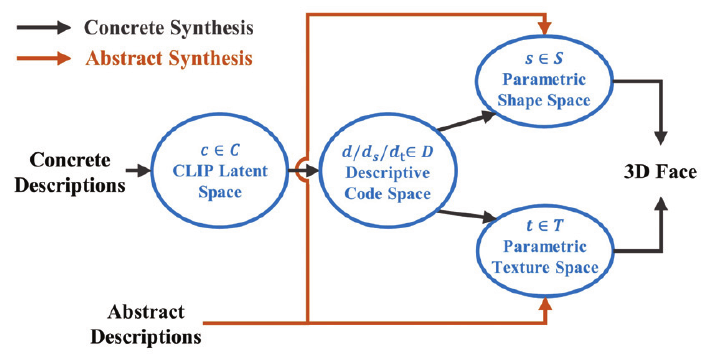}
    \vspace{-0.2in}
    \caption{Relationship of the involved parametric spaces.
    }
    \vspace{-0.1in}
    \label{fig:mapping}
\end{figure}

\begin{figure*}
    \centering
    \includegraphics[width=1.0\linewidth]{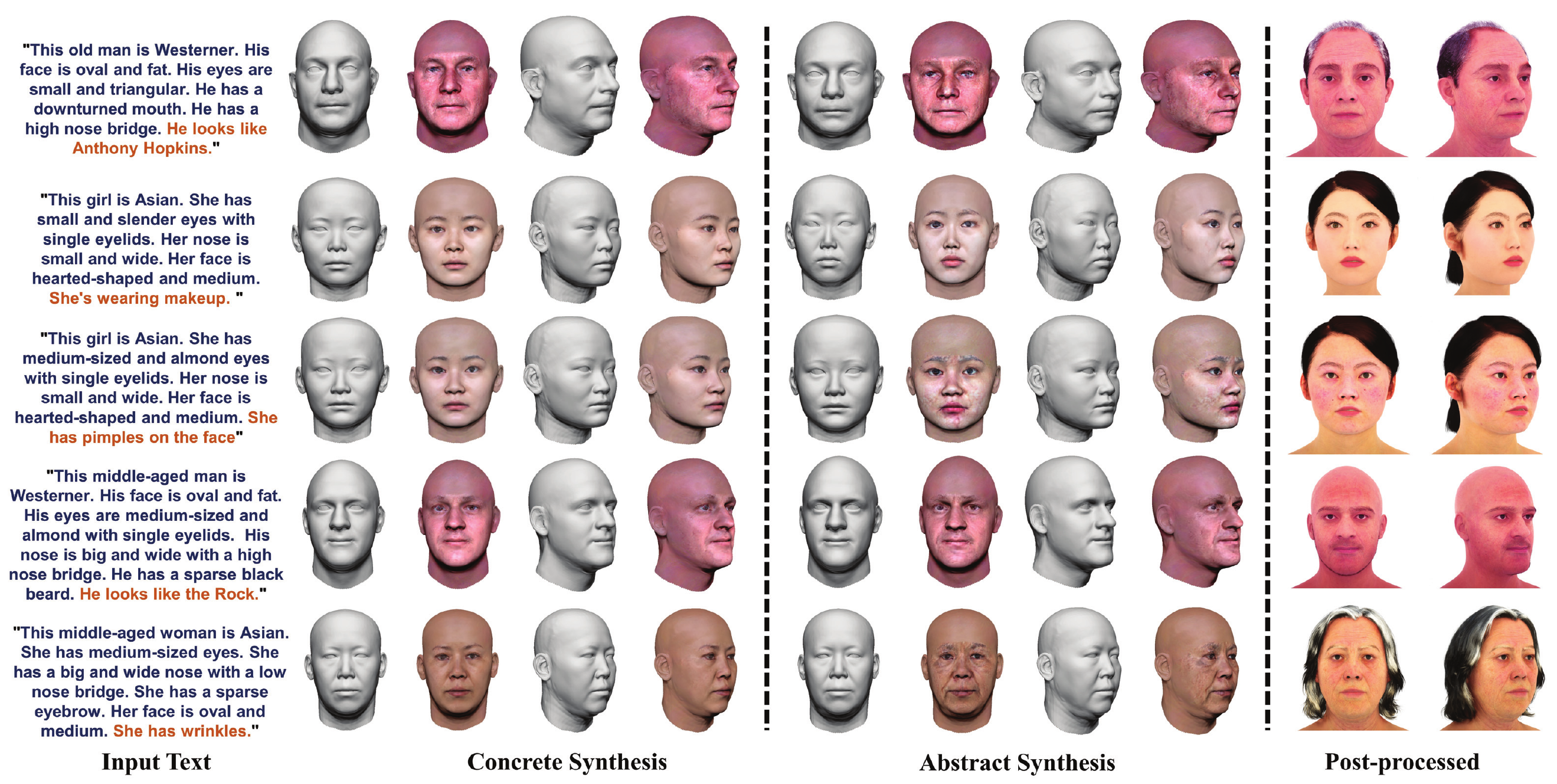}
    \vspace{-0.2in}
    \caption{Qualitative evaluations of our method for text-to-3D face generation. Our pipeline can synthesize 3D faces from concrete (dark blue text) and abstract descriptions (brown text). The hairs and additional accessories can be easily added in the post-process phase.}
    \vspace{-0.1in}
    \label{fig:visual}
\end{figure*}

\subsection{Abstract Synthesis}
\label{sec:abstract}

After a full 3D head model with color is produced from the concrete descriptions, we can further improve the model with the \textit{abstract descriptions} in the input texts, which we refer to as abstract synthesis. 
Abstract descriptions are in free-style describing a certain non-objective characteristic, such as ``looks like Tony Stark'' or ``wearing makeup''.
As shown in Figure~\ref{fig:mapping}, the key idea behind is to leverage prompt learning based on CLIP\cite{radford2021learning}, a large language-vision pre-trained model, to optimize the parameters in $T$ texture space and $S$ 3DMM space. 
Specifically, with the trained and fixed model of the concrete synthesis network, both input abstract descriptions texts and the predicted 3D faces (rendered into image) are encoded into the CLIP latent space. Then the texture parameter $t$ and 3DMM parameter $s$ are optimized to minimize the difference between the predicted 3D face and abstract text descriptions in the CLIP latent space.

Considering that the CLIP model is trained on real-world images, a differentiable renderer is indispensable which renders the generated 3D mesh and UV texture into a portrait. Specifically, we use redner~\cite{li2018differentiable} to render textured mesh as real images at three viewpoints ranging from $-30^{\circ}$ to $+30^{\circ}$ and calculate the cosine similarity between the rendered image and the input prompt. 
The loss function for refining $s$ and $t$ is formulated as:
\begin{align}
\begin{split}
    L_{CLIP} = 1- \left \langle E_T(t),E_I(i) \right \rangle,
\end{split}
\end{align}
where $E_T$ and $E_I$ represent the CLIP text encoder and image encoder respectively, $t$ and $i$ represent the input description and the rendered image, and $\left \langle \cdot,\cdot \right \rangle$ represents the cosine similarity.

We propose to use CLIP Loss to optimize the parameters in the $S$ space and $T$ space generated by the pre-trained model and predict a textured mesh that better matches our prompt description. We set the number of iterations to $200$ by default.

We also add two regularization losses to constrain $S$ space and $T$ space. Our complete loss function is:
\begin{align}
    L_{abstract} = L_{CLIP} + \beta_1\Vert{\hat{s}-s_o}\Vert_2 + \beta_2\Vert{\hat{t}-t_o}\Vert_2,
\end{align}
where $s_o$ and $t_o$ represent the initial value from the concrete synthesis module.
It is worth noting that the abstract synthesis is optional and can be conducted multiple times if more than one prompt text is provided.
We set $\beta_1$=3 and $\beta_2$=0.003 by default.

%% file: 4_exp.tex
\section{Experiments}
\label{sec:exp}

\subsection{Implementation Details}
\label{sec:details}

\noindent \textbf{Training of Text Parser.} 
We randomly generate $1$ million pieces of text descriptions and corresponding descriptive codes $d$ according to our face attribute correspondences, where the text is generated by preset sentence patterns and each text description randomly contains $3$ to all $24$ attributes. The detailed templates and samples will be shown in the supplementary material.
We use the CLIP model to encode the concrete descriptions into $512$-dimensional latent code $c$ and train an 8-layer MLP through a cross-entropy loss to map CLIP code $c$ to descriptive code $d$. 
We use Adam\cite{kingma2014adam} optimizer with a learning rate beginning at 0.001 and decaying after 10 epochs until 20 epochs. We set the batch size to 128.

\noindent \textbf{Training of Shape Generator.} 
We use our \datasetname dataset to form our training sets. We use PCA to convert the model into a $300$-dimensional vector and generate corresponding one-hot code from text annotations to form data pairs. For all data, we randomly select $80\%$ for training and the other $20\%$ for testing. We use weighted $\ell_1$ Loss and RST Loss to train our shape generator, an 8-layer MLP. In the first layer, we concatenate the input one-hot code and a 512-dimensional normally distributed noise into the network to generate diverse results. We use ReLU as our activation function.

\noindent \textbf{Training of Texture Synthesis Networks.} 
We follow the hyper-parameters and training settings of StyleGAN to train the mapping network and texture generator. The resolution of the UV texture maps for training and testing is $512\times512$.

\subsection{Qualitative Evaluation}

We present our main experimental results in Figure~\ref{fig:visual}.
We observe that our proposed method can synthesize 3D faces that exactly match the input concrete descriptions (text in dark blue in \Cref{fig:visual}), then these generated 3D faces can be improved to reflect abstract descriptions (text in brown), including ``look likes Anthony Hopkins'', ``be wearing makeup'', etc.
The hairs and additional accessories can also be easily added via 3D modeling software like MetaHuman Creator~\cite{fang2021metahuman} (right column of \Cref{fig:visual}). More results will be shown in the supplementary material.

\subsection{Comparison Experiments}

\begin{figure}
    \centering
    \includegraphics[width=1.0\linewidth]{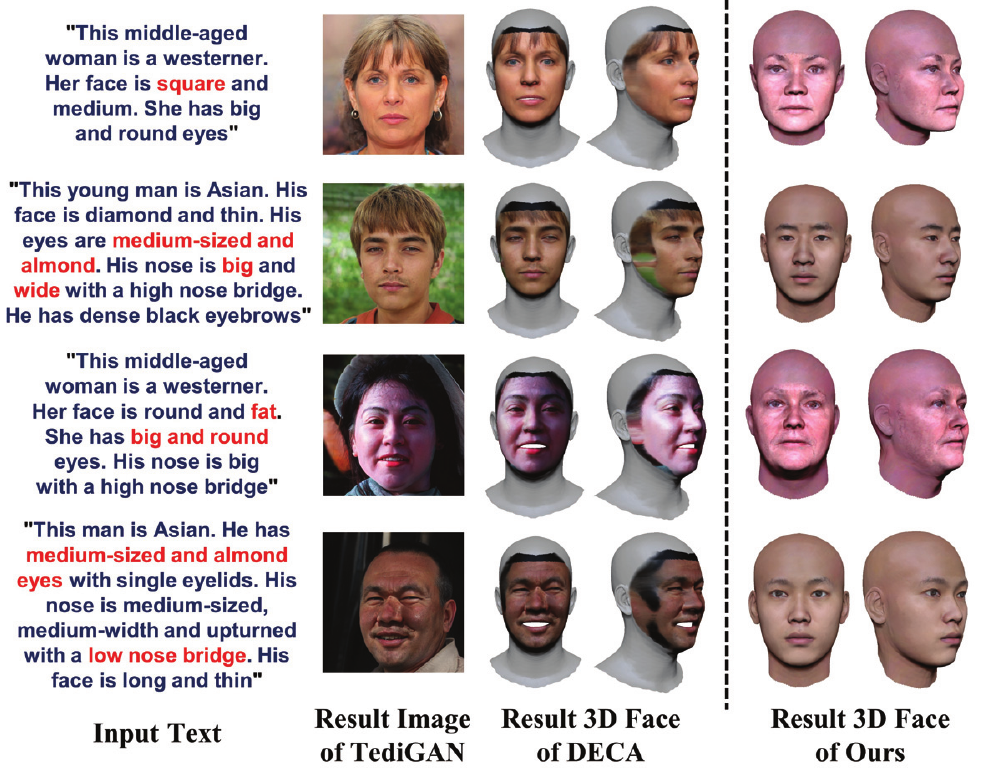}
    \vspace{-0.2in}
    \caption{Comparison with TediGAN\cite{xia2021tedigan}+DECA\cite{feng2021learning}. The red texts indicate the descriptions that the results of TediGAN+DECA do not match.}
    \vspace{-0.1in}
    \label{fig:comp_2d}
\end{figure}

\begin{figure}
    \centering
    \includegraphics[width=1.0\linewidth]{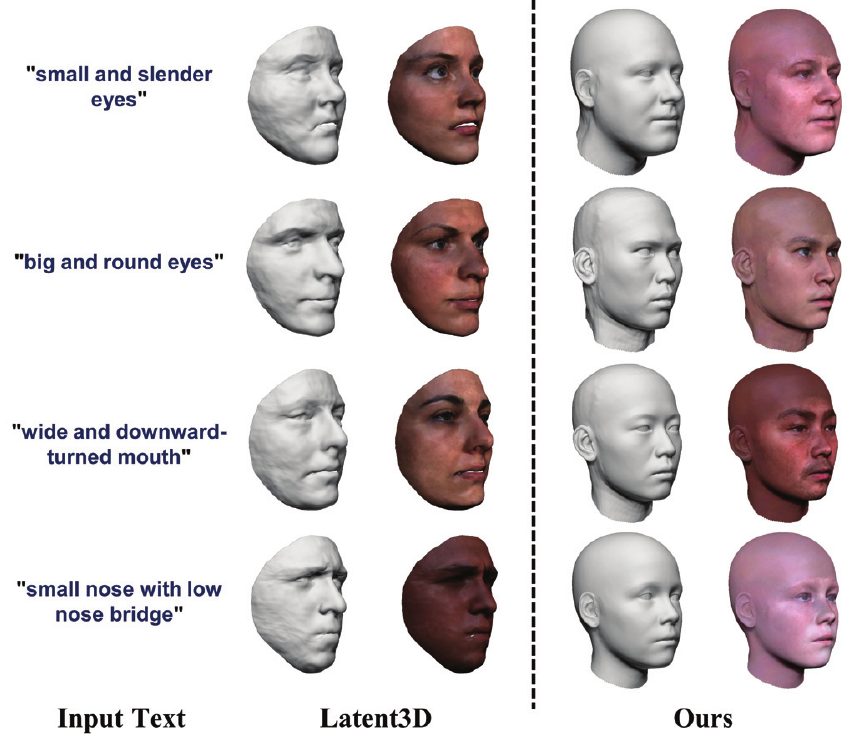}
    \vspace{-0.2in}
    \caption{Comparison with Latent3D~\cite{canfes2022text}.}
    \vspace{-0.15in}
    \label{fig:comp_latent3d}
\end{figure}

\begin{figure*}
    \centering
    \includegraphics[width=0.95\linewidth]{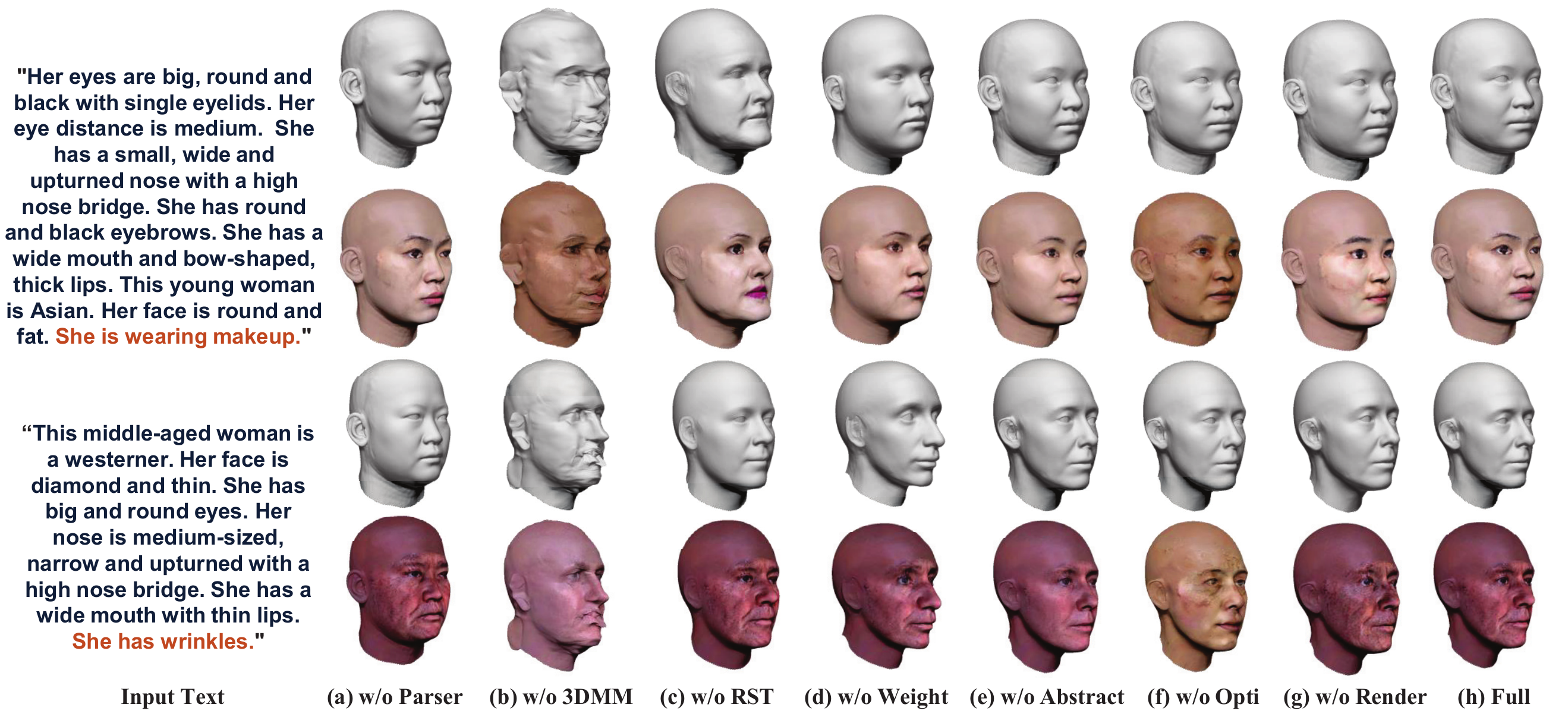}
    \vspace{-0.1in}
    \caption{Generated 3D faces when removing or replacing a certain module in our proposed pipeline for ablation study.}
    \vspace{-0.1in}
    \label{fig:ablation}
\end{figure*}

We compare our method with the two most relative previous works. We use Chamfer Distance (CD), Complete Rate (CR) to measure the accuracy of the generated 3D shape, and use Relative Face Recognition Rate (RFRR) \cite{sun2022anyface} to measure the identity similarity of the textured 3D face. The precise definition of these metrics will be explained in the supplementary material. 

\noindent \textbf{Text-Image-Shape.} 
The task of generating a 3D face from descriptive texts can be achieved by cascading the Text-to-Image model and Image-to-Shape model in an end-to-end manner. Here we choose TediGAN~\cite{xia2021tedigan}, a SOTA Text-to-Image model, and DECA~\cite{feng2021learning}, a SOTA single-view face reconstruction model to compose the Text-to-Shape model. As shown in Figure~\ref{fig:comp_2d}, the Text-Image-Shape strategy fails to match many input descriptions (red texts), and Table~\ref{tab:comp_2d} shows that our method outperforms Text-Image-Shape in all three metrics. We believe this is due to the fact that the TediGAN and DECA cannot be optimized end-to-end. Besides, depth ambiguity commonly exists in the in-the-wild image datasets, which leads to inaccurate shape generation of the Text-Image-Shape strategy.

\noindent\textbf{Latent3D~\cite{canfes2022text}.} Latent3D can synthesize a 3D face using text or image-based prompts. 
As Latent3D can only work for short sentence input, while the performance degraded severely for a long paragraph, we only fed a short descriptive sentence into Latent3D for comparison. 
Besides, Latent3D set a random face as an initial face for refinement, and we select a random seed to generate the initial face in the comparison experiment.
As shown in Figure~\ref{fig:comp_latent3d}, the result generated by Latent3D fail to recover fine-grained facial features like ``round face'' and ``slender eyes''. Besides, Latent3D relies on an initial guess, and the descriptions can not be matched if the initial guess if the initial value deviates too much from the description. By contrast, our method synthesizes a 3D face that conforms to the description and also supports more detailed descriptions as input. The quantitative evaluation shown in Table~\ref{tab:comp_latent3d} also demonstrates the superior performance of our model.

\begin{table}
\centering
\caption{Quantitative comparison with Text-Image-Shape.}
\vspace{-0.1in}
\begin{tabular}{@{}lccc@{}}
\toprule
\multicolumn{1}{c}{Method} & CD ($mm$) $\downarrow$  & CR ($\%$) $\uparrow$     & RFRR $\uparrow$ \\ \midrule
Text-Image-Shape               & 2.78  & 83.9  & 0.471   \\
Ours                           & \textbf{2.26}  & \textbf{96.7} & \textbf{0.788}   \\ \bottomrule
\end{tabular}
\label{tab:comp_2d}
\end{table}

\begin{table}
\centering
\caption{Quantitative comparison with Latent3D (only the front face error is calculated due to the generative form of Latent3D).}
\vspace{-0.1in}
\begin{tabular}{@{}lccc@{}}
\toprule
\multicolumn{1}{c}{Method} & CD ($mm$) $\downarrow$  & CR ($\%$) $\uparrow$     & RFRR $\uparrow$ \\ \midrule
Latent3d~\cite{canfes2022text} & 2.40  & 94.3  & 0.542   \\
Ours                           & \textbf{1.53}   & \textbf{99.1}  & \textbf{0.778}   \\ \bottomrule
\end{tabular}
\vspace{-0.1in}
\label{tab:comp_latent3d}
\end{table}

\subsection{Ablation Study}
\label{sec:ablation}

\noindent \textbf{Effect of Text Parser and Concrete Synthesis.} 
To validate the effectiveness of the introduction of descriptive code, 3DMM representation, RST loss, and the weights for $\ell_1$ loss, we conduct the experiments with the following settings:

\noindent $\bullet$ (a) Without Descriptive Code: The descriptive code is not used and the embedding vector generated from the input text by the CLIP encoder is directly fed into the concrete synthesis module.

\noindent $\bullet$ (b) Without 3DMM: The network of 3DMM parameter regressing is replaced by a position map generator, of which the backbone is StyleGAN2\cite{Karras2019stylegan2}.

\noindent $\bullet$ (c) Without RST loss: The RST loss is removed from the loss function of the shape generation network.

\noindent $\bullet$ (d) Without weights of $\ell_1$ loss: The weights in the $\ell_1$ loss to train the shape generation network are set to $1$.

The visualized results of the ablation study are shown in Figure~\ref{fig:ablation}. We can see that our full method generates a detailed faithful 3D face.
Comparing (a) with (h), we find the results of (a) failed to match the input descriptions, which verified that the introduction of parsed supervision improves the effectiveness of our model. We consider the reason is that the introduction of descriptive code decouples the complex text-to-3D problem into two simpler problems: 1) parsing text to explicitly categorized facial features in the form of one-hot code and 2) generating 3D face from this one-hot code. 
Comparing (b) with (h), we find the directly predicted shape is distorted, which demonstrates that 3DMM based shape generator is superior to a non-parametric generator in our task. 
Comparing (c) and (d) with (h), we find the facial features in (c) and (d) are not obvious, though most of the features match the input description. It demonstrates that the recognition of the resulting facial features is enhanced after the RST loss and the weights for $\ell_1$ loss is implemented.

\noindent \textbf{Effect of Abstract Synthesis.}
To validate the effectiveness of the abstract synthesis, optimization method, and differentiable render, we conduct the experiments with the following settings:

\noindent $\bullet$ (e)  without abstract synthesis: The phase of abstract synthesis is removed. 

\noindent $\bullet$ (f) prompt: optimization $\rightarrow$ train: In the abstract synthesis phase, the strategy to optimize $s$ and $t$ is changed to adding CLIP loss to the loss function and fine-tuning the model of the concrete synthesis network.

\noindent $\bullet$ (g) without differentiable renderer: In the abstract synthesis phase, the differentiable renderer is removed, and only UV texture is updated through CLIP loss.

We draw the following observations.
Comparing (e) with (h), we can see that the abstract descriptions of ``makeup'' and ``wrinkles'' appear in (h) while the other facial features are consistent with (e), which demonstrates the effectiveness of abstract synthesis. 
Comparing (f) with (e), we find the training with clip loss failed to synthesize abstract features while the other facial features in (e) are not maintained. By contrast, the prompt learning strategy in (h) synthesizes more plausible results.
Comparing (g) with (h), we find that our method without differentiable render may hallucinate unnatural features. For example, in the first line, the lipstick color in (g) is painted beyond the lips. We consider the reason is that the CLIP model is trained using real-world images, but the UV texture space is distorted compared to the real space, so a differentiable renderer is a requisite to transform the facial appearance from UV texture space to real-world space.

%% file: 5_con.tex
\vspace{-0.03in}
\section{Conclusion}
\vspace{-0.05in}
\label{sec:con}

In this work, we investigate the problem of generating a 3D face from descriptive texts in natural language. 
To this end, a \textsc{Describe3D} dataset is developed by annotating descriptions to large-scale 3D face datasets.
We first train neural networks to generate a 3D face matching the concrete description and random coding, then optimize the parameters of 3DMM and StyleGAN space with abstract description to further refine the 3D face model. Experiments show that our method can produce a faithful 3D face that conforms to the input description. 

There are some drawbacks to our approach. First, our method requires a pre-distinction between concrete and abstract descriptions, and the performance degrades when the input sentences are significantly different from the template sentences we adopt. Besides, as the number of races in the dataset is not balanced, the modeling effect of the facial features of some ethnic minorities is poor.

\noindent\textbf{Acknowledgement.} This work was supported by the NSFC grant 62001213, 62025108, and gift funding from Huawei and Tencent Rhino-Bird Research Program. Thanks to Shi Zong for his valuable advice on this work.

%% file: 6_supple.tex
\section{Supplementary Materials}

\appendix

\section{Overview}
In the supplementary material, we first explain the details of our \datasetname dataset in Section~\ref{sec:data}, including all the annotated attributes and all descriptive options. Then more implementation details and details about the evaluation metrics are explained in Section~\ref{sec:implement} and Section~\ref{sec:metric}. Finally, more results regarding qualitative revaluation and comparison experiments are presented in Section~\ref{sec:more_results}.

\section{Dataset}
\label{sec:data}

\subsection{3D Models.} 
The 3D models in our \datasetname dataset are collected from HeadSpace~\cite{dai2020statistical} and FaceScape~\cite{yang2020facescape}. The collected model contains a 3D triangle mesh and a UV texture map, and is transformed into the topologically uniform models with Procrustes analysis~\cite{gower1975generalized} and non-rigid iterative closest point (NICP) algorithm~\cite{amberg2007optimal}. The processed topologically uniform mesh model consists of $26369$ vertices and $52536$ triangle faces, attached with a UV texture map at a resolution of $1024\times1024$. The UV texture map is down-sampled into $512\times512$ resolution to train the texture generation network.

\subsection{Descriptive Texts} 
\begin{figure}
    \centering
    \includegraphics[width=1.0\linewidth]{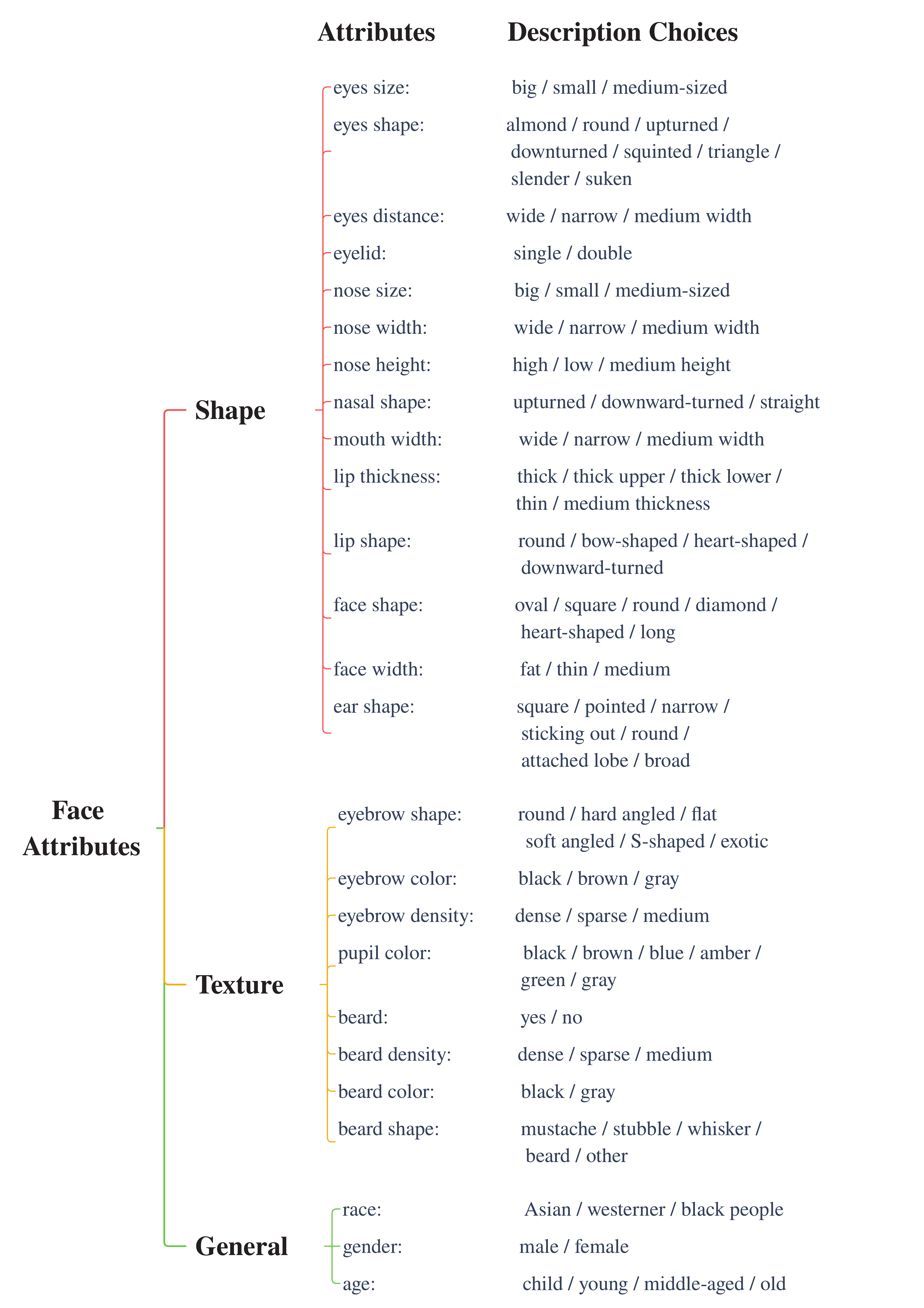}
    \vspace{-0.2in}
    \caption{
    Attributes and descriptive choices in our annotation questionnaire. 
    }
    \label{fig:attribute}
\end{figure}

We established a questionnaire containing $25$ single-choice questions and one short-answer question, and request a professional labeling institute to complete the labeling task with this questionnaire. The high-quality rendered images of the faces in the front view and side view are shown to the annotators. As shown in Figure~\ref{fig:attribute}, each question is about a facial attribute (left column), and each choice represents a description of this attribute (right column). Illustrations about these descriptions are inserted to help annotators understand the meaning of these descriptions. The short-answer question is ``\emph{Please observe the main view and side view of the face picture, describe the facial features of the face of each group of pictures in detail, as far as possible through your description, you can reproduce all the details of the face, and your facial feature description can distinguish the face and other faces.}''. 

\begin{figure}
    \centering
    \includegraphics[width=1.0\linewidth]{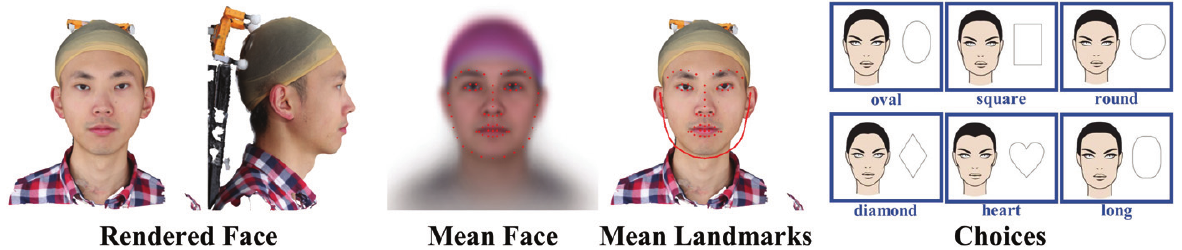}
    \vspace{-0.2in}
    \caption{
    Example of the visualization. 
    }
    \vspace{-0.1in}
    \label{fig:annotation}
\end{figure}

For the convenience of the annotation, we visualize the models as shown in Figure~\ref{fig:annotation}. We first collect the 3D face models from FaceScape and HeadSpace datasets, normalize the scale and position, then render the models at the front and side views. The mean face and the extracted facial landmarks are visualized for judgment. The annotators are required to complete a questionnaire containing 25 multiple-choice questions and a free description according to the rendered images. The secondary labeling and sampling inspection are conducted to ensure the accuracy of the labeling.

\subsection{Text Composer}  
According to the annotated attributes, we use pre-defined sentence patterns to generate text descriptions for each 3D face model. It is worth noting that we use the composed texts but not the answer texts of the short-answer question to train and test our model because the answer texts are not comprehensive enough.

The composed descriptive texts contain 7 sentences to describe eyebrows, eyes, nose, mouth, face shape, race, gender and age, and beard respectively. We pre-defined two sentence patterns. Taking eyes as an example, we use "His eyes are ..." or "He has ... eyes" to form our sentences.
The order and the number of sentences can be adjusted to augment the texts for training, which will be detailed in Section~\ref{sec:text_parser}.

\section{Implement Details}
\label{sec:implement}

\subsection{Training of Text Parser}
\label{sec:text_parser}

To train the text parser, we generate an augmented dataset containing an input text and corresponding descriptive code. Since the input text can be generated from the descriptive code as explained in Section~\ref{sec:data}, we first generate $1,000,000$ random and non-redundant descriptive codes by combining random $24$ attributes (ears are not included), then generate corresponding descriptive sentences. Each sentence contains $1$ or more attributes of a specific region, such as "His face is fat" (1 attribute) and "His face is fat and round" (2 attributes). In each training iteration for a certain region, we randomly select $2-7$ sentences from the composed sentences to generate a training tuple, and the order of the sentences is shuffled.

\begin{figure*}
    \centering
    \includegraphics[width=1.0\linewidth]{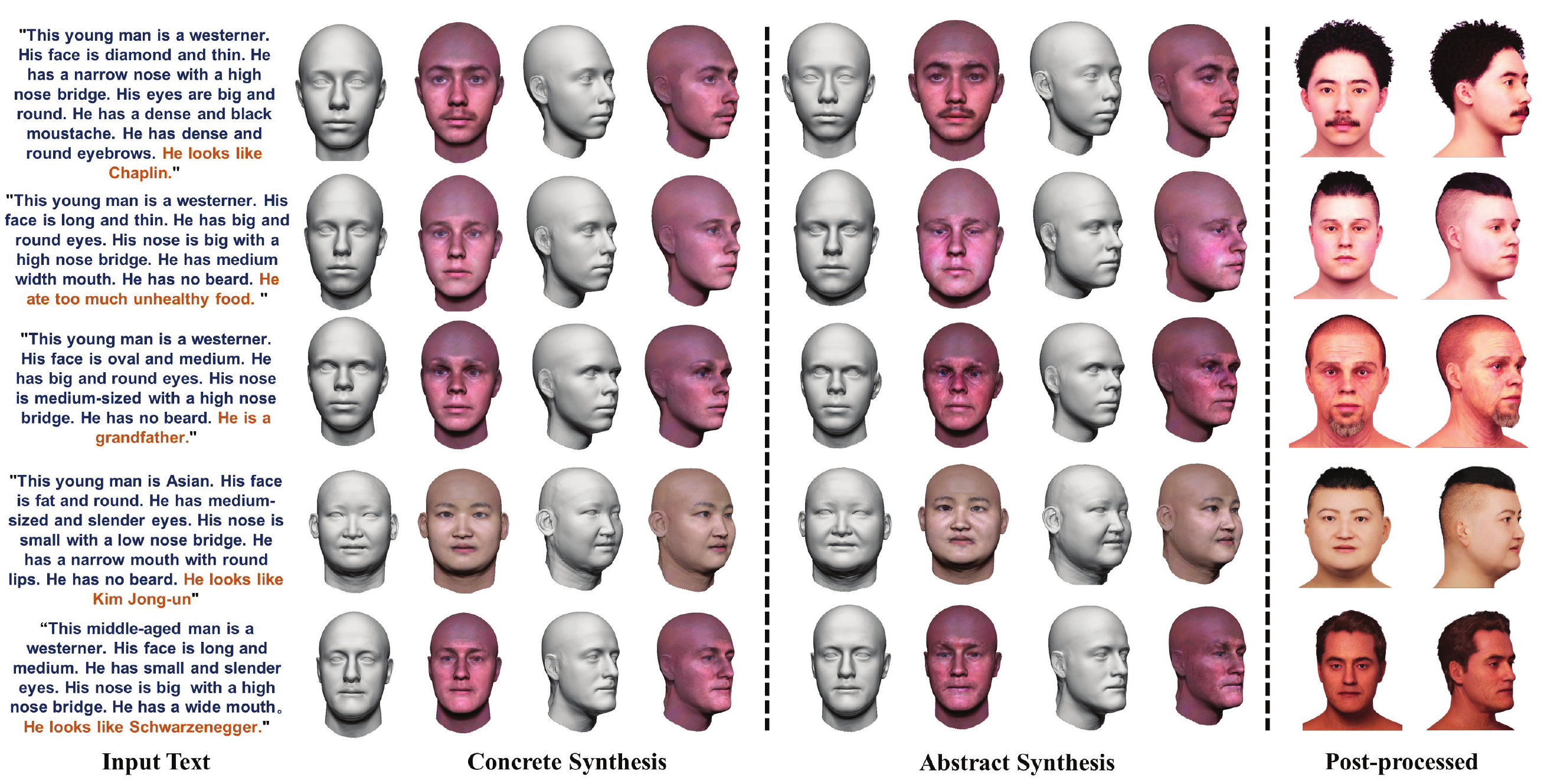}
    \vspace{-0.2in}
    \caption{More qualitative results.}
    \vspace{-0.1in}
    \label{fig:more_qualitative}
\end{figure*}

\subsection{Region-Specific Triplet Loss}

We propose Region-Specific Triplet Loss (RST Loss) to train our shape generator, which is formulated as:

\begin{align}
    {L}_{RST} = \max(\Vert{\hat{v_i} - v_i}\Vert_1 - \Vert \hat{v_i} -  v_i^{*}\Vert_1 + m_i , 0) \cdot {\lambda}_i,
\end{align}where $\hat{v_i}$ is the predicted vertices of $i$-th region, $v_i$ is the corresponding ground-truth, and $v_i^*$ is its counter example. $m_i$ and ${\lambda}_i$ represent corresponding region margin and weight respectively.

Concretely, we choose four regions and randomly select one region for training in each iteration. We pre-establish a mapping list between positive and negative samples, for example, "big and high nose" is the negative sample for "small and low nose", and "fat and round face" is the negative sample for "long and thin face". In the training phase, we randomly select a sample as the positive example, then select the corresponding negative example according to the mapping list and train the model with the RST loss. 
In all our experiments, $r$ is set to the average value of the RST loss between all positive and negative examples, and the threshold $m_i$ is equal to $r$. The weight ${\lambda}_i$ is set to eliminate the influence of different scales in different regions.

\subsection{Post-process}

We use MetaHuman Creator~\cite{metahuman} to automatically register the generated 3D face into a riggable model, then manually add hair and skin texture. 
The 3D shape of the generated model and the post-processed model is highly consistent (mean error distance $< 0.3mm$).
MetaHuman Creator cannot fit hair and skin textures, so hair and skin textures are manually assigned from the assets library, and we think this is the reason for visual inconsistencies between before and after the post-processed. The post-processed 3D faces can be directly used in rigging and animation pipelines.

\section{Evaluation Metric}
\label{sec:metric}

We use three metrics for quantitative evaluation: Chamfer Distance (CD), Complete Rate (CR), and Relative Face Recognition Rate (RFRR). We will explain how they are calculated below.

\noindent $\bullet$ \textbf{Chamfer Distance(CD)}: CD measures the overall error distance.  Given the processed predicted mesh $\mathcal{M}_{p}$ and the ground-truth mesh $\mathcal{M}_{g}$, chamfer distance is formulated as:

\begin{equation}
\begin{aligned}
\begin{split}
{CD}({\mathcal{M}_{p}}, {\mathcal{M}_{g}}) = &\frac{1}{N_{p}} \sum\limits_{x\in{\mathcal{M}_{p}}}^{N_{p}} \sum\limits_{y\in{\mathcal{M}_{g}}}^{N_{g}} {\min \left\| {x - y} \right\|}_2 + \\ &\frac{1}{N_{g}} \sum\limits_{y\in{\mathcal{M}_{g}}}^{N_{g}} \sum\limits_{x\in{\mathcal{M}_{p}}}^{N_{p}} {\min \left\| {x - y} \right\|}_2,
\end{split}
\end{aligned}
\end{equation}
where $N_{p}$, $N_{g}$ are the numbers of the vertices of the predicted mesh and the ground-truth mesh respectively. Since Latent3D~\cite{canfes2022text} only reconstructs the front face, we extract the front face from our predicted mesh and then calculate the CD.

\noindent $\bullet$ \textbf{Complete Rate(CR)}: CR measures the integrity of the reconstruction results, which is formulated as:
\begin{equation}
\begin{aligned}
\eta = \frac{P_1}{P_0},
\end{aligned}
\end{equation}
where $P_1$ is the number of points with a CD value less than $10mm$ and $P_0$ is the number of all points.

\noindent $\bullet$ \textbf{Relative Face Recognition Rate(RFRR)}:
Since there is no general standard for measuring 3D face texture, we choose to use the Relative Face Recognition Rate(RFRR) similar to that in Anyface~\cite{sun2022anyface}. We render the predicted mesh and ground-truth mesh with the same camera parameters and use ArcFace~\cite{deng2019arcface} to extract features that represent facial identities. Then we calculate the cosine similarity and use it to measure the similarity between the ground-truth face and the predicted face.

We align the predicted 3D model to the ground-truth model before the computations of metrics. Specifically, we scale the predicted model to match the scale of the ground-truth model to have a consistent interpupillary distance. Then, Iterative Closest Point (ICP) algorithm is applied to align the predicted mesh to the ground-truth mesh. 

\section{More Results}
\label{sec:more_results}

\subsection{More Visual Results and Comparisons}
\begin{figure}
    \centering
    \includegraphics[width=1.0\linewidth]{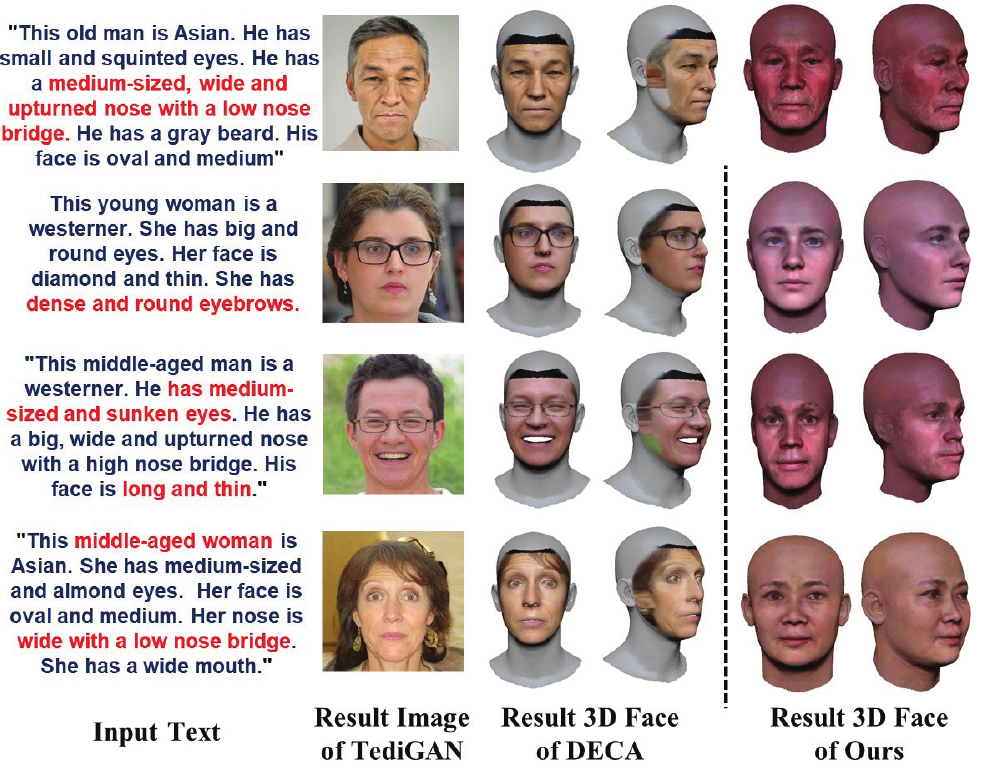}
    \vspace{-0.2in}
    \caption{
    More comparisons to TediGAN~\cite{xia2021tedigan}+DECA~\cite{feng2021learning}. 
    }
    \vspace{-0.1in}
    \label{fig:more_comp_2d}
\end{figure}

\begin{figure}
    \centering
    \includegraphics[width=1.0\linewidth]{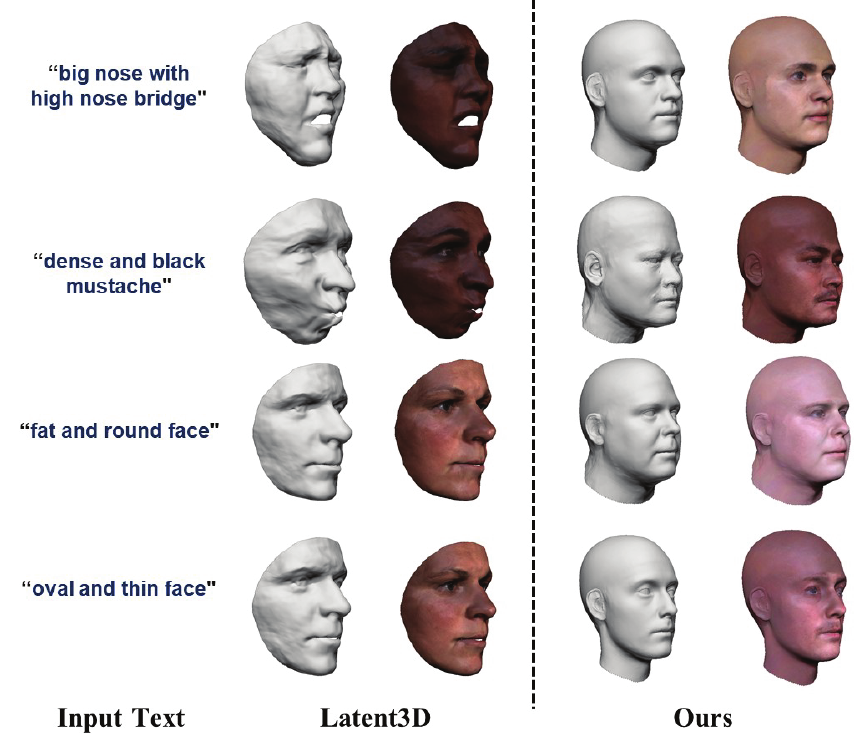}
    \vspace{-0.2in}
    \caption{
    More comparisons to Latent3d. 
    }
    \vspace{-0.1in}
    \label{fig:more_comp_latent3d}
\end{figure}

We show more qualitative results in Figure ~\ref{fig:more_qualitative}, which is the extension of Figure 6 in the main paper. We also show more comparison results in Figure~\ref{fig:more_comp_2d} and Figure~\ref{fig:more_comp_latent3d}, corresponding to Figure 7 and Figure 8 in the main paper.

\subsection{Diverse Results}
\label{sec:diverse_result}

\begin{figure}[t]
    \centering
    \includegraphics[width=1.0\linewidth]{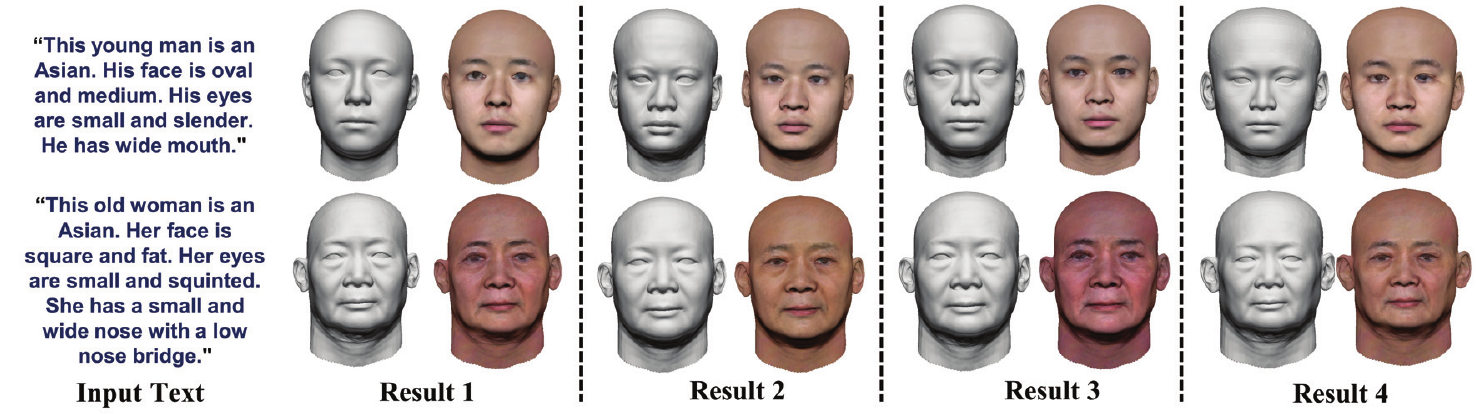}
    \vspace{-0.2in}
    \caption{Diverse results generated from different noise. 
    }
    \vspace{-0.2in}
    \label{fig:noise}
\end{figure}

We add an extra noise vector in the shape generation network. This design is based on the fact that a given descriptive text can correspond to many diverse 3D faces. Therefore, we add a noise vector as input to increase the diversity of the generation, and the effectiveness is verified in Fig~\ref{fig:noise}. As we model the shape generation with the 3DMM regressing problem, the adversarial loss is not involved since it is not
suitable for a parameter-regressing network.

\subsection{Failure Cases}

\begin{figure}
    \centering
    \includegraphics[width=1.0\linewidth]{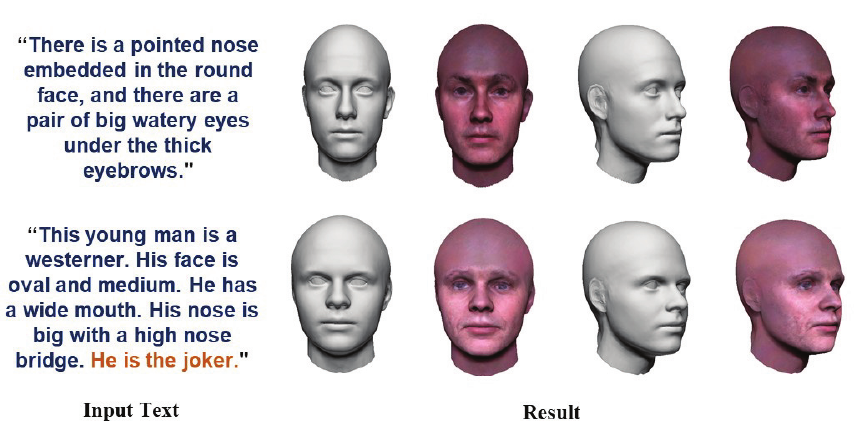}
    \vspace{-0.2in}
    \caption{
    Failure cases. 
    }
    \vspace{-0.1in}
    \label{fig:failure}
\end{figure}

We found that our approach produced some failures, which can be categorized into two categories:

\noindent \textbf{Casual descriptive texts.} As shown in the first row of Figure~\ref{fig:failure}, our model may fail with casual descriptive texts, which contains complex sentence patterns like ``pointed nose embedded in ...'', and figurative description like ``big watery eyes''. In this case, it is obvious that the result 3D face doesn't match the input description of ``round face''. We think the main reason is that our model is trained with relatively simple sentence patterns, and there is still room to improve the generalization of the text parser.

\noindent \textbf{Special appearance.} The shape and appearance of our results are strictly constrained in the $S$ and $T$ spaces that are built upon the training set, therefore, the abstract synthesis stage cannot generate a face with a non-human appearance. As shown in the second row of Figure~\ref{fig:failure}, given ``joker'' as prompt, the abstract synthesis can generate a wide mouth which is a typical feature of the jokers in the films. However, the other features like the red nose and exaggerated clown makeup can not be generated, since these features are not covered in our $S$ and $T$ space.

